\documentclass[a4paper,fleqn]{cas-sc}

\usepackage[authoryear,longnamesfirst]{natbib}
\usepackage{bbding}
\usepackage{siunitx}

\def\tsc#1{\csdef{#1}{\textsc{\lowercase{#1}}\xspace}}
\tsc{WGM}
\tsc{QE}
\begin{document}
\let\WriteBookmarks\relax
\def\floatpagepagefraction{1}
\def\textpagefraction{.001}

% Short title
\shorttitle{}    

% Short author
\shortauthors{P. Pavlík et al.}

% Main title of the paper
\title [mode = title]{The Utility of Physics-Informed Machine Learning in Solar Power Production Forecasting}  

% Title footnote mark
% eg: \tnotemark[1]
% \tnotemark[1] 

% Title footnote 1.
% eg: \tnotetext[1]{Title footnote text}
% \tnotetext[1]{} 

% First author
%
% Options: Use if required
% eg: \author[1,3]{Author Name}[type=editor,
%       style=chinese,
%       auid=000,
%       bioid=1,
%       prefix=Sir,
%       orcid=0000-0000-0000-0000,
%       facebook=<facebook id>,
%       twitter=<twitter id>,
%       linkedin=<linkedin id>,
%       gplus=<gplus id>]

\author[1,2]{P. Pavlík}[orcid=0000-0002-7468-5503]
\cormark[1]
\cortext[1]{Corresponding author}
\ead{peter.pavlik@kinit.sk}
\credit{Conceptualization, Methodology, Software, Data Curation, Formal analysis, Investigation, Visualization, Writing – original draft}

% Address/affiliation
\affiliation[1]{organization={Kempelen Institute of Intelligent Technologies},
            addressline={Sky Park Offices, Bottova 7939/2A}, 
            city={Bratislava},
            postcode={811 09},
            country={Slovakia}}

% Address/affiliation
\affiliation[2]{organization={Faculty of Information Technology, Brno University of Technology},
            addressline={Božetěchova 2}, 
            city={Brno},
            postcode={612 00},
            country={Czech Republic}}

\author[3]{Franz M. Rohrhofer}[orcid=0000-0002-8228-7759]

% Credit authorship
\credit{Supervision}

% Address/affiliation
\affiliation[3]{organization={Know Center Research GmbH},
            addressline={Sandgasse 34}, 
            city={Graz},
            postcode={8010},
            country={Austria}}

\author[3,4,5]{Bernhard C. Geiger}[orcid=0000-0003-3257-743X]

% Credit authorship
\credit{Project administration, Supervision, Writing - Review \& Editing}

% Address/affiliation
\affiliation[4]{organization={Signal Processing and Speech Communication Laboratory, Graz University of Technology},
            addressline={Inffeldgasse 16c}, 
            city={Graz},
            postcode={8010},
            country={Austria}}

% Address/affiliation
\affiliation[5]{organization={Research Center ENERGETIC, Graz University of Technology},
            addressline={Rechbauerstraße 12}, 
            city={Graz},
            postcode={8010},
            country={Austria}}

% For a title note without a number/mark
%\nonumnote{}

% Here goes the abstract
\begin{abstract}
Physics-informed machine learning (PIML) and hybrid modeling have attracted significant attention as approaches for combining physical knowledge with the flexibility of deep learning. In this work, we investigate whether these approaches deliver their commonly claimed benefits for large-scale photovoltaic (PV) power forecasting. We develop two hybrid forecasting architectures that integrate the ADR physical PV efficiency model with the TSMixer neural forecasting architecture through feature augmentation and residual learning. The proposed models are evaluated on a dataset comprising over 3 million samples from 3,713 PV installations across 33 European countries. We assess convergence speed, forecasting accuracy, transferability across geographical regions, and data efficiency. Although the physics-informed models seem to exhibit slightly faster convergence during training, they do not achieve consistent improvements in test-set accuracy, generalization, or performance under limited training data compared to the purely data-driven baseline. These findings suggest that, for modern deep forecasting architectures trained on large datasets, explicitly incorporating the ADR physical model provides limited practical benefit for day-ahead PV forecasting.
\end{abstract}

% Use if graphical abstract is present
%\begin{graphicalabstract}
%\includegraphics{}
%\end{graphicalabstract}

% Research highlights
% \begin{highlights}
% \item Developed two PIML architectures that integrate a PV efficiency model with TSMixer.
% \item Performed one of the largest evaluations of physics-informed PV forecasting to date.
% \item Evaluated the commonly claimed advantages of PIML and found no clear benefits.
% \end{highlights}

% Keywords
% Each keyword is seperated by \sep

\maketitle

\section{Introduction} \label{sec:intro}

The growing transition toward renewable energy sources has led to a rapid increase in the share of solar power within the global energy mix. However, photovoltaic (PV) systems are inherently dependent on weather conditions, which remain highly variable and difficult to predict. Sudden fluctuations in solar power generation can introduce instability into the electrical grid and, if unaccounted for, result in blackouts~\cite{HASSAN2026116324}. Accurate forecasting of PV power output is therefore essential, enabling grid operators to better manage supply-demand balance by dispatching stored energy or activating backup generation sources.

Machine learning (ML) approaches have proven highly effective for solar power forecasting, typically formulating the problem as a time series prediction task using historical production data and weather forecasts. Modern architectures, such as mixer-based models, can achieve high predictive accuracy while capturing site-specific characteristics and complex temporal dependencies~\cite{CAO2026112554, en18040765}. By leveraging the historical data, these approaches can easily outperform methods based solely on physical inputs such as solar irradiance forecasts.

At the same time, physics-informed machine learning (PIML) and hybrid modeling have emerged as a promising paradigms for integrating physical knowledge with data-driven models~\cite{karniadakis2021physics}. By incorporating domain-specific constraints or differentiable physical components, PIML methods are widely reported to improve overall accuracy of the model, robustness under unseen conditions, reduce physically implausible predictions, and increase model interpretability~\cite{karniadakis2021physics, willard2022integrating, cuomo2022scientific}. While similar benefits have also been reported in the context of PV forecasting~\cite{s22030749, POMBO20226512, WANG2025136546}, prior work has primarily focused on incorporating physics through engineered features or auxiliary constraints in relatively small-scale studies~\cite{WANG2026127378}. The integration of explicit, differentiable PV performance models as components within end-to-end forecasting architectures remains underexplored.

In this work, we investigate whether such physics-informed integration provides measurable benefits in large-scale PV forecasting. We develop a hybrid physics-informed forecasting framework by embedding a differentiable ADR model of PV module efficiency into a TSMixer neural forecasting architecture. The resulting hybrid model is trained end-to-end and evaluated on a large-scale dataset comprising thousands of PV installations across Europe. The unmodified TSMixer architecture serves as the baseline, with the comparison focusing on the task of deterministic day-ahead forecasting.

Our evaluation considers predictive accuracy, generalization and transferability across sites, and data efficiency. Contrary to commonly reported advantages of PIML, our results do not show consistent improvements over the baseline, nor evidence of faster convergence. We hypothesize that this outcome is due to two factors: (i) the limited fidelity of the employed physical PV efficiency model, which may constrain the usefulness of the imposed physical structure, and (ii) the strong representational capacity of modern architectures like TSMixer, which are able to implicitly learn relevant physical relationships from data without explicit guidance.

These findings suggest that the benefits of physics-informed approaches in PV forecasting may be more context-dependent than initially assumed, particularly in high-data regimes with advanced models. At the same time, the proposed framework provides a practical example of integrating differentiable physical models into forecasting architectures and establishes a foundation for future work exploring alternative physical formulations and hybrid modeling strategies.

\section{Related Work}

Recent work has explored the integration of physical knowledge into ML models for PV power forecasting, typically under the framework of PIML. A common approach is to incorporate physical relationships through engineered features derived from analytical PV models. For example, \cite{s22030749} propose augmenting standard meteorological inputs with physics-informed features that capture the operational state of PV systems, such as data transformations based on irradiance and estimated temperature. Their claim is that while the ML model, given enough iterations, should be capable to learn these relationships by itself, by computing them in advance they can speed up the learning process, reduce memory needs and improve accuracy. Their results demonstrate that such features can consistently improve forecasting performance across a range of ML models. Similarly, in a follow-up benchmarking study~\cite{POMBO20226512}, the authors evaluate multiple ML approaches and show that models enriched with physics-informed features tend to outperform their purely data-driven counterparts. Notably, all of the tested approaches rely on feature engineering rather than integrating physical models directly into the learning process.

More recent studies have investigated tighter integration of physical knowledge using physics-informed neural networks (PINNs). For instance, \cite{WANG2025136546} model PV module temperature using an ordinary differential equation embedded within a PINN framework. The predicted temperature is then used as an additional input to a downstream forecasting model, leading to improved predictive performance. Notably, their approach assumes access to ground truth module temperature for training and validation, which may not be available in many real-world scenarios. In contrast, we aimed to develop a model that does not rely on direct measurements of module temperature and instead incorporates a simplified efficiency model based solely on commonly available inputs.

Another line of work focuses on embedding physical constraints directly into the training objective. \cite{WANG2025126764} propose a dynamic-parameter PINN framework in which physical equations governing PV conversion efficiency are incorporated via a composite loss function. Their method introduces adaptive physical parameters and weather-dependent weighting, enabling improved performance under extreme conditions. While this approach demonstrates strong gains, it is evaluated on data from a single large-scale PV installation, raising questions about its generalizability across diverse sites and operating conditions.

In parallel to these developments, purely data-driven approaches have achieved strong performance in PV forecasting without explicit physical modeling. Recurrent neural networks, particularly long short-term memory (LSTM) models, have been widely adopted due to their ability to capture temporal dependencies in historical production and weather data. More recently, attention-based architectures and mixer-style models have demonstrated further improvements by modeling long-range dependencies and complex feature interactions more effectively than recurrent approaches.

Recent works have demonstrated the strong performance of mixer-based architectures for PV power forecasting, particularly the TSMixer model~\cite{ekambaram2023tsmixer}. For example, \cite{CAO2026112554} propose a hybrid framework combining TSMixer with a two-stage signal decomposition and residual correction strategy, further enhanced by Bayesian hyperparameter optimization. Their results, validated across multiple PV installations in diverse climatic regions of Australia and China, show that TSMixer-based models can achieve high accuracy and robustness. Similarly, \cite{en18040765} employ TSMixer for transfer learning with dynamic time warping (DTW) to improve forecasting performance in small-scale PV systems with limited data. Their study, conducted on multiple installations in the United Kingdom, demonstrates that TSMixer consistently outperforms both LSTM- and transformer-based models in terms of prediction accuracy and reliability.

These results highlight the strong representational capacity and flexibility of modern data-driven architectures, particularly TSMixer-based models, which are capable of implicitly learning relationships between meteorological variables and PV output when trained on sufficiently large datasets. As such, they reduce the need for manual feature engineering or explicit physical constraints and represent a competitive and modern baseline for PV forecasting.

Overall, existing PIML approaches to PV forecasting differ primarily in how physical knowledge is incorporated: as expert-engineered input features, auxiliary predicted variables, or loss-based constraints. In contrast, the integration of differentiable physical models as explicit components within end-to-end forecasting architectures remains relatively underexplored. Moreover, most prior studies evaluate their methods on data from a single or a small number of PV installations, limiting the ability to assess generalization across heterogeneous systems.

In this work, we address these gaps by embedding a differentiable PV efficiency model directly within a neural forecasting architecture and evaluate its impact at scale across thousands of PV installations distributed across Europe. This setup allows us to systematically assess whether the commonly claimed benefits of physics-informed approaches hold in a large-scale, high-data regime with a strong modern baseline.

\section{Physical Model of PV Efficiency} \label{sec:ADR}

To incorporate physical knowledge into the forecasting model, we employ the ADR PV efficiency model~\cite{9323039}. This model provides a compact analytical description of PV module efficiency as a function of irradiance and cell temperature, using a small number of physically interpretable parameters. It should serve as a good basis for the proposed PIML approach.

Before applying the efficiency model, the incident irradiance on the panel surface is computed from standard meteorological inputs. Specifically, global, direct, and diffuse horizontal irradiance (GHI, DNI, DHI) are transformed into plane-of-array (POA) irradiance using panel geometry and solar position. The direct component is projected onto the tilted panel surface based on the angle of incidence, while diffuse irradiance is decomposed into sky and ground-reflected components. The sky diffuse contribution is approximated using an isotropic model, and the ground-reflected component depends on surface albedo and panel tilt. The total POA irradiance is then obtained as the sum of direct and diffuse contributions and serves as the effective irradiance input ($S$) to the ADR model.

At a high level, the ADR model captures how PV efficiency changes with operating conditions relative to standard test conditions. The input irradiance is first normalized with respect to a reference value (1000 W/m²), and the deviation of cell temperature from a reference temperature (25°C) is computed. These normalized quantities are then used to model the electrical behavior of the PV module, particularly the relationship between irradiance, voltage, and efficiency.

Specifically, the ADR model describes PV module efficiency as a function of normalized irradiance $S$ and cell temperature $T$ using a set of analytical expressions with five parameters ($k_a, k_d, tc_d, k_{rs}, k_{rsh}$).
The model is computed in four steps. First, a temperature-dependent “dark irradiance” term is defined as
\begin{equation}
S_o(T) = 10^{k_d + tc_d (T - 25^\circ\mathrm{C})}.
\end{equation}
Using this quantity, the voltage-like term is expressed as
\begin{equation}
V(S, T) = \ln\left(\frac{S}{S_o(T)} + 1\right),
\end{equation}
which captures the nonlinear dependence of PV voltage on irradiance and temperature.
Next, the voltage is normalized with respect to reference conditions:
\begin{equation}
    v(S, T) = \frac{V(S, T)}{V(1, 25^\circ\mathrm{C})}. \label{eq:voltage}
\end{equation}
Finally, the module efficiency is given by
\begin{equation}
    \eta(S,T) = k_a [(1 + k_{rs} + k_{rsh}) v(S,T)-k_{rs}S-k_{rsh}v(S,T)^2].
\end{equation}

The parameters have clear physical interpretations. The scaling factor $k_a$ represents the efficiency at reference conditions. The parameter $k_d$ controls the irradiance–voltage relationship through the dark irradiance term, while $tc_d$ introduces temperature dependence into this relationship. The coefficients $k_{rs}$ and $k_{rsh}$ represent series and shunt resistance losses, respectively, and can be interpreted as fractional power losses at reference conditions.

This formulation yields a differentiable, closed-form approximation of PV efficiency that captures the dominant dependencies on irradiance and temperature while remaining computationally efficient.

In practice, direct measurements of PV cell temperature are rarely available in large-scale datasets. Therefore, we estimate the operating temperature using the Faiman model~\cite{Faiman}, an empirical thermal model widely adopted in PV performance standards. The Faiman model assumes a steady-state energy balance between absorbed solar irradiance and heat dissipation to the environment. The module temperature is given by

\begin{equation}
T_{\text{cell}} = T_{\text{air}} + \frac{G}{u_0 + u_1 \cdot v_{\text{wind}}},
\end{equation}
where $G$ is the incident irradiance, $T_{\text{air}}$ is the ambient temperature, $v_{\text{wind}}$ is the wind speed, and $u_0, u_1$ are empirical heat loss coefficients.

\subsection{Fitting the physical parameters}  \label{sec:ADR_fit}

While the ADR and Faiman models provide reasonable default parameter values, accurate modeling of PV systems requires adapting these parameters to individual installations. In large-scale datasets, however, the amount and quality of data vary significantly across sites, making fully independent parameter estimation prone to overfitting for data-scarce installations.

To address this, we adopt a hybrid global–local parameterization of the physical model. Specifically, each model parameter is decomposed into a shared global component and a site-specific deviation:
\begin{equation}
\theta_i^{s} = \theta_i^{\text{global}} + \Delta \theta_i^{s},
\end{equation}
where $\theta_i^{s}$ denotes the parameter $i$ for site $s$, $\theta_i^{\text{global}}$ is a learnable global parameter, and $\Delta \theta_i^{(s)}$ is a site-specific offset obtained from a learned embedding.

This formulation allows the model to share statistical strength across all installations while still capturing local variations in system behavior. Sites with limited data naturally rely more on the global parameters, whereas sites with sufficient data can learn meaningful deviations. In preliminary experiments, this hybrid approach consistently outperformed both fully shared (global-only) and fully independent (site-specific) parameterizations in terms of overall predictive accuracy.

The site-specific deviations are implemented using a learnable embedding indexed by installation, producing a vector of parameter offsets for each site. To ensure physically meaningful parameter ranges and stable training, appropriate transformations are applied: for example, efficiency-related parameters are constrained to be positive using softplus functions, bounded parameters are mapped via sigmoid or hyperbolic tangent functions, and scaling factors are adjusted to match expected physical ranges.

This parameterization enables end-to-end learning of both global and local physical characteristics within the differentiable modeling framework, making it suitable for large-scale heterogeneous datasets like the one we are using.

\section{Dataset}

\begin{figure}
    \centering
    \includegraphics[width=0.8\columnwidth]{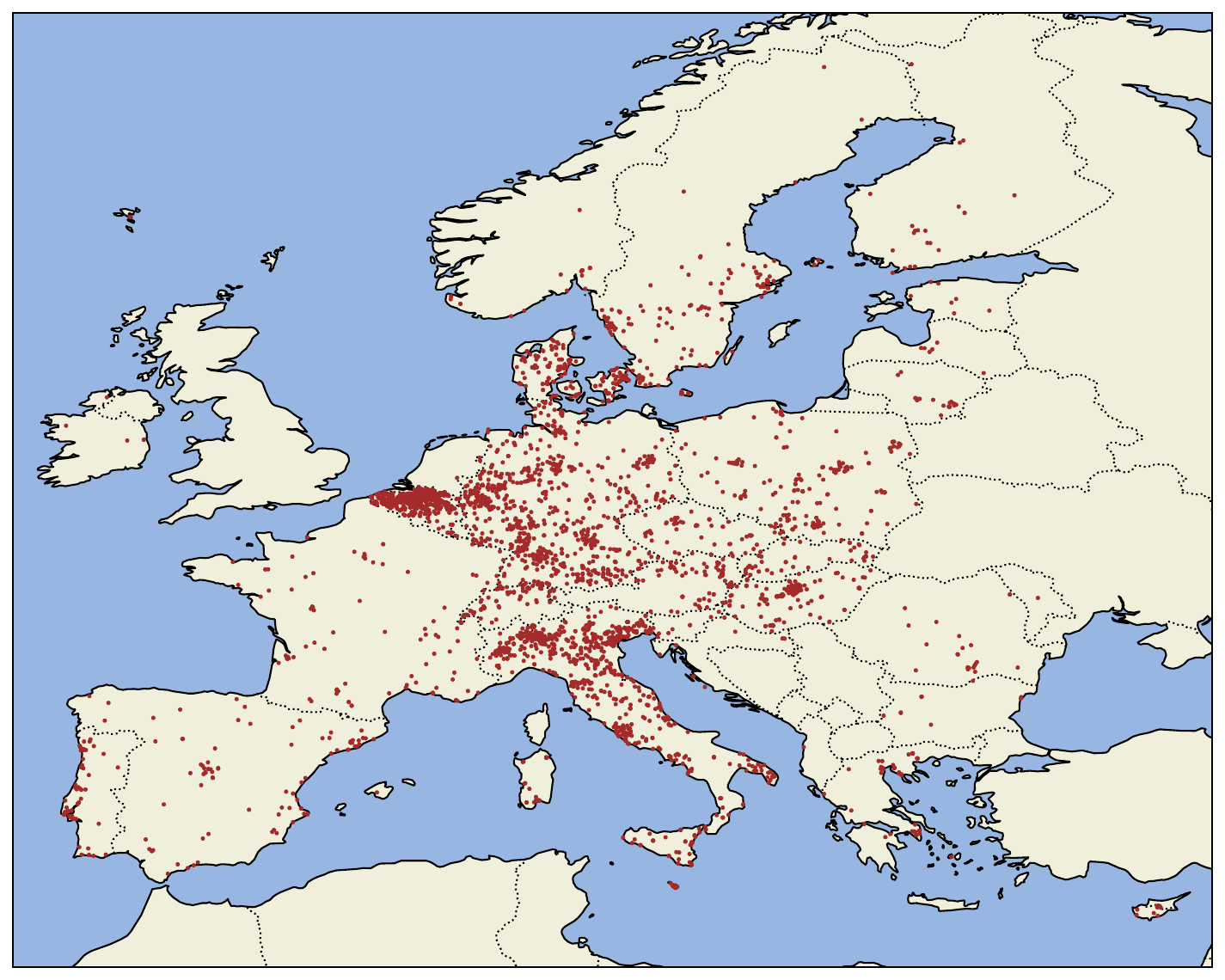}
    \caption{The map shows the geographical distribution of PV installations in the used dataset. Each dot represents one site with at least one day of production data available.}
    \label{fig:installations}
\end{figure}

The dataset used in this study combines PV power production data with corresponding numerical weather prediction (NWP) forecasts across a large number of installations. It contains daily time series from 3,713 PV systems located in 33 European countries, spanning the period from 2020 to 2023. In total, the dataset comprises 3,082,656 samples of daily data. For the geographical distribution of the installation sites, see Figure~\ref{fig:installations}.

The PV production data were obtained from PVOutput.org~\cite{pvoutput_general}, a publicly available platform collecting generation data from distributed solar installations. Weather data were sourced from Deutscher Wetterdienst (DWD) and consist of forecasts generated by the ICON numerical weather prediction (NWP) model~\cite{dwd_iconeu_g5dm}. For each day, the input features correspond to the most recent day-ahead forecast, ensuring a realistic forecasting setup.

Each daily sample includes the observed daily energy production in 15-minute intervals along with meteorological variables such as irradiance, temperature, and wind speed from NWP linearly interpolated to the same time interval. While additional atmospheric variables (e.g., cloud cover, humidity, precipitation) are available, all models in this work use only a subset of inputs corresponding to the physical model requirements: global and direct irradiance, ambient temperature, and wind speed.

In addition to time-varying features, the dataset includes installation-specific metadata such as panel tilt and orientation, which are used in the physical modeling pipeline to compute POA irradiance.

The large scale and geographic diversity of the dataset introduce substantial variability in weather conditions, system configurations, and data quality. This makes it a suitable benchmark for evaluating both generalization and transferability of forecasting models across heterogeneous PV installations.

\section{Methodology}

The objective of this study is to evaluate whether incorporating physical knowledge into a modern deep learning forecasting model provides measurable benefits for day-ahead PV power forecasting. Specifically, we investigate several commonly claimed advantages of PIML, including improved forecasting accuracy, faster convergence during training, reduced data requirements, and improved generalization across heterogeneous PV installations.

Unlike previous work, our objective is not to develop a more sophisticated physical model, but rather to investigate whether an existing differentiable physical model can improve the performance of a state-of-the-art forecasting architecture when evaluated on a large and diverse real-world dataset.

\subsection{Models}

To answer this question, we compare four models representing different levels of integration between physical modeling and data-driven learning: a purely physical model, a purely data-driven model, and two physics-informed hybrid models. All models produce deterministic one-day-ahead forecasts of normalized daily PV production, where values are scaled to the interval $[0,1]$ by the installed capacity of the corresponding PV system.

The ADR model described in Section~\ref{sec:ADR} serves as the purely physical baseline. The model receives numerical weather prediction inputs, computes the POA irradiance and module temperature, and produces a deterministic estimate of the normalized daily PV production. All ADR and Faiman parameters are learned end-to-end using the differentiable framework described in Section~\ref{sec:ADR}, employing the hybrid global--local parameterization introduced in Section~\ref{sec:ADR_fit}.

As the purely data-driven baseline, we employ TSMixer~\cite{ekambaram2023tsmixer} using the implementation provided in the NeuralForecast library~\cite{olivares2022library_neuralforecast}. TSMixer was selected because recent work has demonstrated state-of-the-art performance for PV forecasting while maintaining a comparatively simple architecture and efficient training procedure. The model receives the same meteorological inputs as the ADR model but additionally uses a five-day history of both PV production and weather variables, allowing it to exploit temporal dependencies that are unavailable to the physical model. Throughout the remainder of this paper, this model is referred to as \textit{TSMixer-Baseline}.

The first hybrid model, denoted \textit{TSMixer-ADRAugment}, extends the baseline TSMixer by augmenting its input with physically meaningful quantities computed by the ADR pipeline. Besides the final ADR prediction, these include intermediate variables such as the estimated module temperature deviation from reference conditions, normalized irradiance, normalized voltage (Eq.~\eqref{eq:voltage}), POA direct and diffuse irradiance, and the solar incidence projection factor. The network architecture and training procedure remain otherwise identical to those of the baseline model. Within the taxonomy of PIML, this approach incorporates physical knowledge through engineered features derived from an analytical model, similarly to the methodology proposed by Pombo et al.~\cite{s22030749}.

The second hybrid model, referred to as \textit{TSMixer-ADRResidual}, uses the same augmented feature set but adopts a residual learning strategy. Instead of directly predicting the normalized PV production, the network learns the difference between the ADR prediction and the observed production. The final forecast is obtained by adding the predicted residual to the ADR estimate. This formulation encourages the neural network to model only those effects that are not captured by the physical model, while relying on ADR to explain the dominant irradiance--temperature relationship. Furthermore, the residual target exhibits substantially lower variance than the original production values, which may facilitate optimization and accelerate convergence during training.

Table~\ref{tab:model_comparison} summarizes the information available to each model.

\begin{table}
\centering
\caption{Comparison of evaluated forecasting models' inputs.}
\label{tab:model_comparison}
\begin{tabular}{lccc}
\hline
Model & Weather & History & ADR Features \\
\hline
ADR & \Checkmark & \XSolidBrush & -- \\
TSMixer-Baseline & \Checkmark & \Checkmark & \XSolidBrush \\
TSMixer-ADRAugment & \Checkmark & \Checkmark & \Checkmark \\
TSMixer-ADRResidual & \Checkmark & \Checkmark & \Checkmark \\
\hline
\end{tabular}
\end{table}

Together, these four models enable a systematic evaluation of the contribution of physical knowledge at different levels of integration. Comparing ADR with TSMixer-Baseline quantifies the benefits of purely data-driven learning, while comparisons between the hybrid models and TSMixer-Baseline isolate the effect of incorporating physically derived information into an otherwise identical forecasting architecture.

\subsection{Experimental Setup}

The experimental evaluation is organized into three complementary studies, each designed to investigate one of the primary research questions posed in this work. The first experiment evaluates forecasting accuracy and convergence speed during training to determine whether incorporating physical knowledge improves predictive performance or accelerates optimization. The second experiment investigates transferability by evaluating model performance across countries with substantially different amounts of available training data. Finally, the third experiment examines data efficiency by progressively reducing the amount of available training data, allowing us to assess whether the proposed physics-informed approaches maintain their performance under data-scarce conditions.

All experiments employ the same data partitioning strategy. The dataset is split chronologically, with the final 20\% of samples reserved for testing and the remaining 80\% used for training. From the training partition, 1\% of the samples are randomly selected to form a validation set used for monitoring the training process and model selection.

All neural network models are trained by minimizing the mean squared error (MSE) using the Adam optimizer with an exponentially decaying learning rate. Hyperparameters were selected based on preliminary experiments and kept fixed throughout all evaluations. To ensure a fair comparison, each physics-informed model uses the same neural network architecture, optimization procedure, and hyperparameter configuration as its corresponding data-driven baseline. Complete model configurations and training hyperparameters are available in the accompanying source code repository.

Although the proposed framework allows joint optimization of both the physical model and the neural network, preliminary experiments showed that optimizing both components from random initialization substantially slowed convergence. Therefore, the ADR model is first calibrated in a separate pre-training stage to provide a good initialization of the physical parameters. During subsequent neural network training, these parameters remain trainable and are further refined jointly with the forecasting model through backpropagation.

Model performance is evaluated using both normalized and absolute error metrics. Since all the models predict normalized PV production, we first report the mean squared error (MSE) between the predicted and ground-truth normalized production values,

\begin{equation}
\mathrm{MSE}=\frac{1}{N}\sum_{i=1}^{N}(\hat{y}_i-y_i)^2,
\end{equation}

where $y_i\in[0,1]$ denotes the normalized PV production and $\hat{y}_i$ the corresponding prediction. In addition to the per-observation MSE, we also evaluate the error after aggregating all observations belonging to the same day,

\begin{equation}
\mathrm{MSE}_{\mathrm{day}}=\frac{1}{D}\sum_{d=1}^{D}\left(\sum_{t=1}^{T_d}\hat{y}_{d,t}-\sum_{t=1}^{T_d}y_{d,t}\right)^2,
\end{equation}

where $T_d$ denotes the number of observations within day $d$.

To assess forecasting accuracy in physically meaningful units, the normalized predictions are converted back to power production using the installed capacity $C$ of the corresponding PV installation. We then compute the mean absolute error (MAE),

\begin{equation}
\mathrm{MAE}=\frac{1}{N}\sum_{i=1}^{N}\left|C\hat{y}_i-C y_i\right|,
\end{equation}

which is reported in watts (W). Likewise, we report the MAE after aggregating the predicted and observed production over an entire day,

\begin{equation}
\mathrm{MAE}_{\mathrm{day}}=\frac{1}{D}\sum_{d=1}^{D}\left|C\sum_{t=1}^{T_d}\hat{y}_{d,t}-C\sum_{t=1}^{T_d}y_{d,t}\right|,
\end{equation}

which is converted to daily energy production and reported in kilowatt-hours (kWh). Unlike the normalized metrics, the denormalized MAE metrics naturally assign greater weight to larger PV installations and can better reflect practical forecasting performance.

The detailed protocols for the three experiments -- \textit{Accuracy and Convergence Evaluation}, \textit{Transferability Evaluation}, and \textit{Data Efficiency Evaluation} -- are described Section~\ref{sec:ACE}, \ref{sec:TE}, and \ref{sec:DEE}, respectively.

\section{Accuracy and Convergence Evaluation} \label{sec:ACE}

This experiment evaluates the primary hypothesis of this work, namely whether incorporating physical knowledge into a modern forecasting architecture leads to improved predictive accuracy or faster convergence during training. We therefore compare the three TSMixer-based models introduced in the previous section: the purely data-driven \textit{TSMixer-Baseline}, the feature-augmented \textit{TSMixer-ADRAugment}, and the residual-learning \textit{TSMixer-ADRResidual}.

To account for the stochastic nature of neural network training, each model is trained ten times using different random seeds, resulting in different parameter initializations and data batching order. Every training run is performed for 100 epochs (of a 1000 batches) without early stopping. After each epoch, the model is evaluated on the validation set, and the checkpoint achieving the lowest validation loss is selected for the final evaluation on the test set.

In addition to the final forecasting accuracy, we analyze the evolution of the validation loss throughout training. This allows us to investigate whether the physics-informed models exhibit faster convergence or more stable optimization, as is frequently claimed in the PIML literature.

Both \textit{TSMixer-ADRAugment} and \textit{TSMixer-ADRResidual} are initialized using the pre-calibrated ADR model described in Section~\ref{sec:ADR_fit}. Rather than keeping the physical model fixed, its parameters remain trainable and are jointly optimized with the neural network throughout training. This enables the hybrid models to adapt the physical model to better complement the neural network while starting from an already favorable initialization.

Although the standalone ADR model serves as the physical baseline throughout this work, it is omitted from the comparisons in this section. Preliminary evaluation showed that its forecasting error is substantially higher than that of all three TSMixer-based models, causing the corresponding plots to become difficult to interpret. The standalone ADR results are revisited in Section~\ref{sec:DEE}, where they provide a useful reference point in the data-efficiency analysis.

\subsection{Results}

\begin{figure}
    \centering
    \includegraphics[width=0.8\columnwidth]{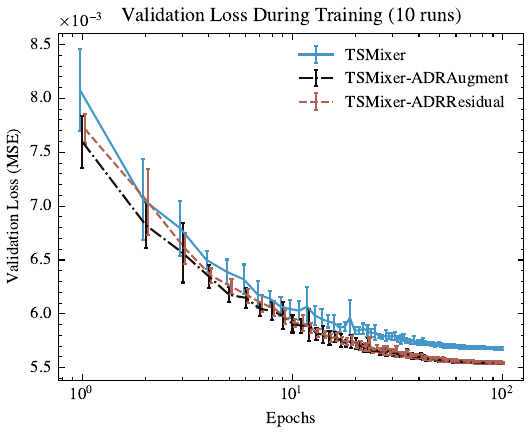}
    \caption{Validation loss (MSE) during training for the evaluated models, averaged over ten independent training runs for each. Lines represent the mean validation loss at each training epoch, while the error bars indicate one standard deviation of the loss across the 10 runs. The horizontal axis is shown on a logarithmic scale to emphasize the convergence behavior during the initial stages of training.}
    \label{fig:validation}
\end{figure}

We begin the evaluation by examining the convergence behavior of the three models during training. For each of the ten training runs per architecture, we record the validation MSE after every training epoch. Figure~\ref{fig:validation} shows the mean validation loss across the ten runs together with their standard deviation.

Figure~\ref{fig:validation} shows that both physics-informed models consistently achieve a lower validation loss than the purely data-driven baseline throughout the entire training process. The difference is already noticeable during the initial training epochs, indicating faster convergence of both \textit{TSMixer-ADRAugment} and \textit{TSMixer-ADRResidual}. As training progresses, all three models gradually converge, although the two PIML variants maintain a small but consistent advantage over the baseline. Furthermore, the relatively small standard deviations suggest that the observed convergence behavior is stable and highly consistent across different random initializations.

The two physics-informed approaches exhibit nearly identical convergence trajectories, with no clear difference in either convergence speed or final validation loss. The addition of the pre-calibrated ADR model therefore appears to facilitate optimization irrespective of whether it is incorporated through feature augmentation or residual learning.

Although these observations suggest that incorporating physical knowledge improves optimization dynamics, validation performance alone is insufficient to assess the overall quality of the forecasting models. We therefore proceed by evaluating their predictive performance on the held-out test set.

\begin{figure}
    \makebox[0.49\columnwidth][l]{\textbf{(a)}}%
    \makebox[0.49\columnwidth][l]{\textbf{(b)}}\\
    \includegraphics[width=0.49\columnwidth]{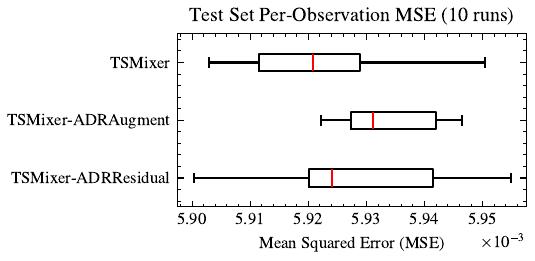}
    \includegraphics[width=0.49\columnwidth]{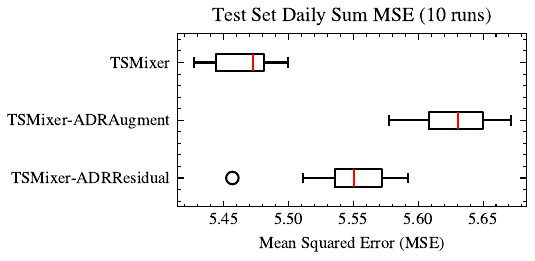}
    \makebox[0.49\columnwidth][l]{\textbf{(c)}}%
    \makebox[0.49\columnwidth][l]{\textbf{(d)}} \\
    \includegraphics[width=0.49\columnwidth]{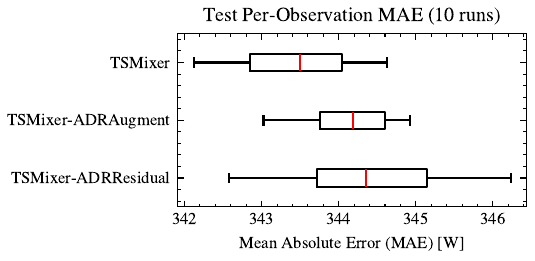}
    \includegraphics[width=0.49\columnwidth]{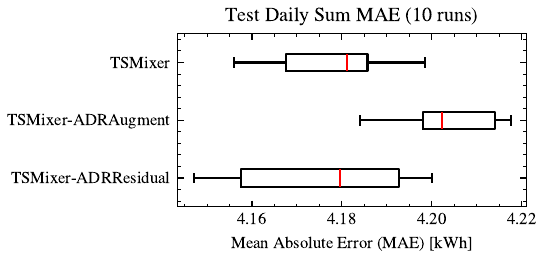}
    \caption{Performance of the evaluated models across ten independent training runs. (a) Per-observation normalized MSE, (b) total daily normalized MSE, (c) per-observation denormalized MAE (W), and (d) total daily denormalized MAE (kWh). Red horizontal lines indicate medians, boxes show the interquartile range (IQR), and whiskers extend to the minimum and maximum observed values (excluding outliers beyond 1.5-times the inter-quartile range).}
    \label{fig:10errors}
\end{figure}

We take the best performing checkpoint (based on validation loss) of each run and compute multiple different metrics for each one. We visualize the results using boxplots for each metric. The plots are in Figure~\ref{fig:10errors}.

The boxplots exhibit inconsistent behavior. Despite the improved validation performance observed for the PIML models, this advantage does not translate into measurable improvements on the held-out test set. While small differences between the models can be observed for some metrics, it is important to consider the scales of the plots. The absolute differences are negligible, indicating that all three models achieve essentially identical predictive performance.

This result suggests that the improved performance observed during training does not necessarily lead to better generalization. A likely explanation is that the physical information provided by the ADR model facilitates optimization by providing a meaningful initialization and physically motivated auxiliary features, but contributes little additional information beyond what the TSMixer is already able to learn directly from the meteorological inputs with enough available data.

\section{Transferability Evaluation} \label{sec:TE}

One of the frequently claimed advantages of PIML is improved generalization under distribution shifts. Since the incorporated physical knowledge is expected to remain valid across different installations and geographical regions, PIML models should, in principle, transfer to different locations more effectively than purely data-driven approaches. To evaluate this claim, we designed an experiment assessing the transferability of the proposed models across different European countries.

The evaluation is based on three Central European countries with markedly different amounts of available training data. Germany serves as a data-rich country with 801 PV installations, while Czechia and Slovakia represent data-scarce settings with only 55 and 47 installations, respectively.

Two complementary transfer scenarios are considered. In the first, models are trained exclusively on the German subset and evaluated on the Czech and Slovak test sets, examining whether knowledge learned from a large and diverse dataset transfers effectively to data-scarce regions. In the second scenario, models are trained on one of the two data-scarce countries and evaluated on the other, representing a somewhat more challenging transfer task where both the source and target domains contain limited training data.

Since the results of the previous experiment showed only marginal differences between \textit{TSMixer-ADRAugment} and \textit{TSMixer-ADRResidual}, we use only the latter as the representative physics-informed approach for a more straightforward comparison. Furthermore, because the ten independent runs with different random seeds converged to nearly identical performance with very little variance, we evaluate only a single training run for each transfer scenario. The models are trained for 100 epochs and the checkpoint with the lowest validation error is selected for evaluation on the target country's test set.

\subsection{Results}

\begin{figure*}
    \makebox[0.49\columnwidth][l]{\textbf{(a)}}%
    \makebox[0.49\columnwidth][l]{\textbf{(b)}}\\
    \includegraphics[width=0.49\columnwidth]{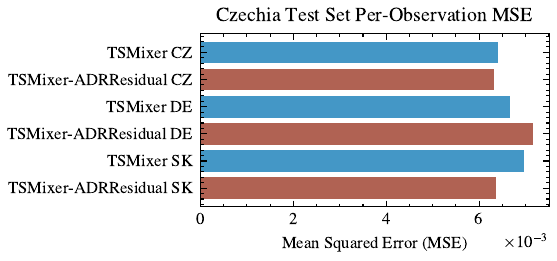}
    \includegraphics[width=0.49\columnwidth]{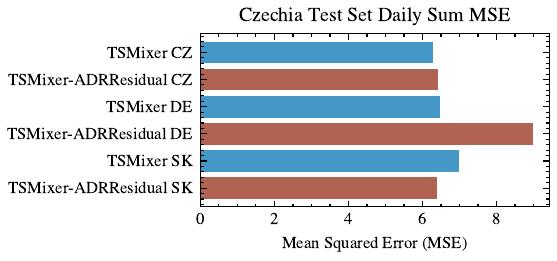}
    \makebox[0.49\columnwidth][l]{\textbf{(c)}}%
    \makebox[0.49\columnwidth][l]{\textbf{(d)}}\\
    \includegraphics[width=0.49\columnwidth]{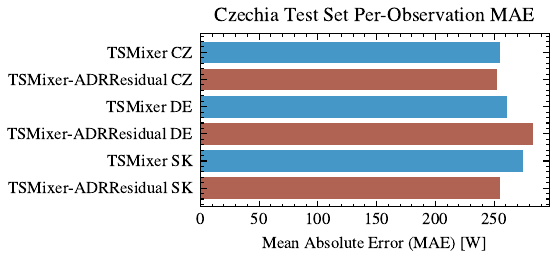}
    \includegraphics[width=0.49\columnwidth]{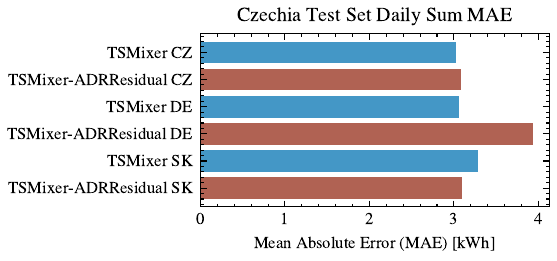}
    \caption{Performance of the evaluated models trained on data from different countries on Czechia test set. (a) Per-observation normalized MSE, (b) total daily normalized MSE, (c) per-observation denormalized MAE (W), and (d) total daily denormalized MAE (kWh). The country code in the label indicates a country whose train data was used to train the model (DE - Germany, CZ - Czechia, SK - Slovakia). Blue bars show the performance of the baseline TSMixer models while the brown bars show the physics-informed TSMixer-ADRResidual models.}
    \label{fig:CZ}
\end{figure*}

We first evaluate the transfer performance on the Czech test set. For each source country, we compare the purely data-driven \textit{TSMixer} and the physics-informed \textit{TSMixer-ADRResidual}, resulting in six evaluated models. Performance is reported using the same four metrics as in the previous experiment: normalized MSE and denormalized MAE, both at the individual 15-minute observation level and after aggregation to daily production totals. The results are shown in Figure~\ref{fig:CZ}.

Figure~\ref{fig:CZ} shows that the models trained on Czech data achieve good overall performance, as expected since the training and test distributions coincide. The two architectures perform almost identically in this setting, indicating that incorporating the ADR model provides little to no benefit here.

When transferring models trained on data from other countries, the models generally exhibit a slight degradation in performance. Interestingly, the German-trained baseline consistently matches or outperforms its physics-informed counterpart across the evaluated metrics, while the Slovak models exhibits the opposite behavior, with TSMixer-ADRResidual performing better. The contrast is even more pronounced for the daily aggregated metrics, where the German \textit{TSMixer-ADRResidual} exhibits the largest prediction errors. The differences are inconsistent and do not indicate a systematic improvement in transferability.

\begin{figure*}
    \makebox[0.49\columnwidth][l]{\textbf{(a)}}%
    \makebox[0.49\columnwidth][l]{\textbf{(b)}}\\
    \includegraphics[width=0.49\columnwidth]{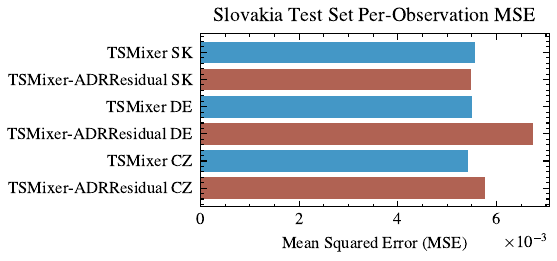}
    \includegraphics[width=0.49\columnwidth]{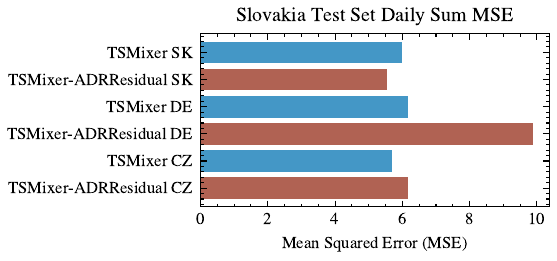}
    \makebox[0.49\columnwidth][l]{\textbf{(c)}}%
    \makebox[0.49\columnwidth][l]{\textbf{(d)}}\\
    \includegraphics[width=0.49\columnwidth]{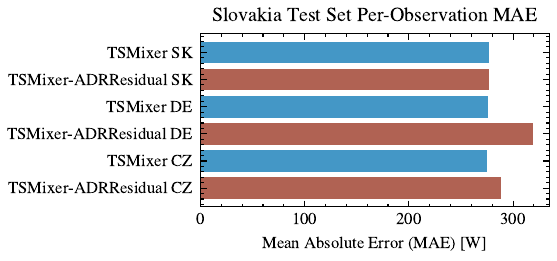}
    \includegraphics[width=0.49\columnwidth]{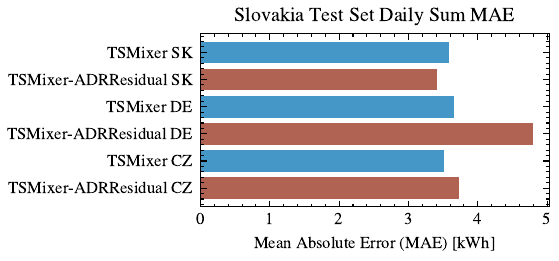}
    \caption{Performance of the evaluated models trained on data from different countries on Slovakia test set. (a) Per-observation normalized MSE, (b) total daily normalized MSE, (c) per-observation denormalized MAE (W), and (d) total daily denormalized MAE (kWh). The country code in the label indicates a country whose train data was used to train the model (DE - Germany, CZ - Czechia, SK - Slovakia). Blue bars show the performance of the baseline TSMixer models while the brown bars show the physics-informed TSMixer-ADRResidual models.}
    \label{fig:SK}
\end{figure*}

We next evaluate the same models on the Slovak test set, with the results shown in Figure~\ref{fig:SK}. Similarly to the Czech evaluation, both locally trained models achieve nearly identical performance. For the transferred models, however, the trend becomes more consistent: both the German- and Czech-trained \textit{TSMixer-ADRResidual} models perform slightly worse than their purely data-driven counterparts across most evaluation metrics. The difference is particularly apparent for the daily aggregated metrics, where the German-trained physics-informed model exhibits the largest prediction errors of all evaluated models.

Taken together, these experiments provide no evidence that incorporating the ADR model improves transferability across countries. Contrary to the common expectation that physics-informed models should generalize better under distribution shifts, the proposed hybrid approach performs similarly to or, in some cases, much worse than the purely data-driven baseline. One possible explanation is that the ADR efficiency model, despite capturing the dominant physical relationships governing PV production, does not adequately represent the diversity of PV technologies, installation geometries, and local operating conditions. Consequently, the physical inductive bias may even introduce a small systematic bias that the neural network must compensate for.

\section{Data Efficiency Evaluation} \label{sec:DEE}

Another commonly cited advantage of PIML is improved sample efficiency, i.e., the ability to achieve competitive predictive performance using substantially less training data. By embedding prior physical knowledge into the learning process, PIML models are expected to rely less on purely statistical patterns and therefore generalize better when only limited observations are available. While the previous section partially addressed this hypothesis by comparing transfer between data-rich and data-scarce countries, the amount of available data and the distribution shift between countries were intertwined. In this experiment, we isolate the effect of training data volume by evaluating all models on samples from the same geographical distribution while systematically varying only the number of training samples.

We compare the \textit{TSMixer} and \textit{TSMixer-ADRResidual} models trained on the complete training dataset and on subsets containing $\frac{1}{10}$, $\frac{1}{100}$, $\frac{1}{1000}$, and $\frac{1}{10000}$ of the original training data. The subsets are sampled from the complete multi-country dataset. For each training subset, the ADR model is first calibrated using the corresponding subset before being used to initialize the PIML model.

All models are evaluated on the same full test set. If the incorporation of physical knowledge indeed improves sample efficiency, the performance gap between the physics-informed and purely data-driven models should become more pronounced as the amount of training data decreases.

\subsection{Results}

In addition to the results of \textit{TSMixer} and \textit{TSMixer-ADRResidual}, we also include the standalone ADR physical model in the comparison. This provides useful context for interpreting the behavior of the physics-informed approach, as it allows us to distinguish between limitations of the physical model itself and limitations of its integration with the neural network.

\begin{figure*}
    \makebox[0.49\columnwidth][l]{\textbf{(a)}}%
    \makebox[0.49\columnwidth][l]{\textbf{(b)}}\\
    \includegraphics[width=0.49\columnwidth]{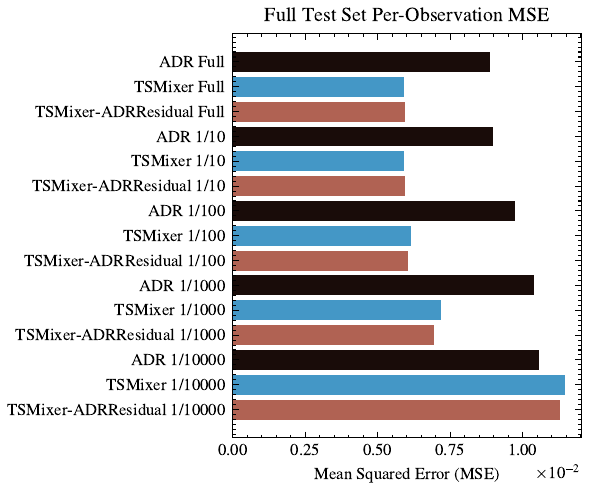}
    \includegraphics[width=0.49\columnwidth]{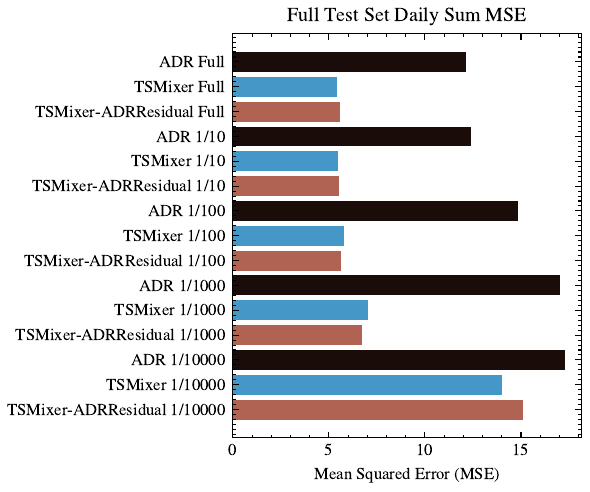}
    \makebox[0.49\columnwidth][l]{\textbf{(c)}}%
    \makebox[0.49\columnwidth][l]{\textbf{(d)}}\\
    \includegraphics[width=0.49\columnwidth]{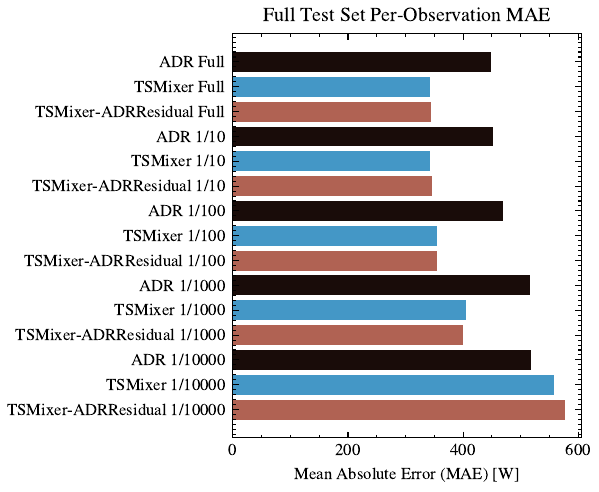}
    \includegraphics[width=0.49\columnwidth]{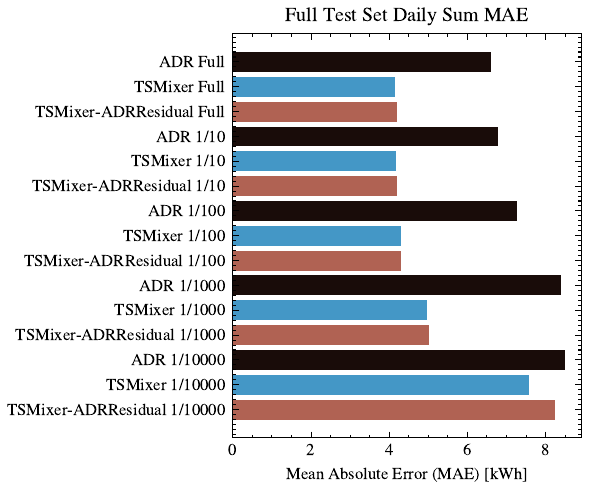}
    \caption{Performance of the evaluated models trained on different fractions of the training dataset. (a) Per-observation normalized MSE, (b) total daily normalized MSE, (c) per-observation denormalized MAE (W), and (d) total daily denormalized MAE (kWh). For each dataset fraction, the standalone ADR model, baseline TSMixer, and physics-informed TSMixer-ADRResidual are compared.}
    \label{fig:Fractions}
\end{figure*}

In total, we compare fifteen models: the ADR baseline, \textit{TSMixer}, and \textit{TSMixer-ADRResidual} trained on the full dataset and on subsets containing $\frac{1}{10}$, $\frac{1}{100}$, $\frac{1}{1000}$, and $\frac{1}{10000}$ of the original training data. As in the previous experiments, performance is evaluated using four metrics: normalized MSE and denormalized MAE, both at the individual 15-minute observation level and after aggregation into daily production totals. The results are shown in Figure~\ref{fig:Fractions}.

The results show that both \textit{TSMixer} and \textit{TSMixer-ADRResidual} are quite robust to reductions in training data. Their performance remains almost unchanged when trained on only $\frac{1}{100}$ of the original dataset, after which the prediction error begins to increase noticeably. The largest degradation occurs for the $\frac{1}{10000}$ subset, where the performance drops rapidly.

The standalone ADR model exhibits a slightly different behavior. Although it is consistently less accurate than the neural approaches when abundant training data are available, its performance deteriorates more gradually as the training dataset becomes smaller. Because ADR relies on a fixed physical formulation rather than purely statistical learning, reducing the amount of training data only affects parameter estimation which already begins at a reasonable starting point. Consequently, at the $\frac{1}{10000}$ subset, the ADR model becomes competitive with the neural models and even achieves lower error in the per-observation metrics.

This observation provides an important insight into the behavior of the proposed physics-informed approach. The ADR model itself clearly retains useful predictive information in data-constrained settings. If the incorporated physical knowledge provided a meaningful inductive bias, one would therefore expect the performance gap between \textit{TSMixer} and \textit{TSMixer-ADRResidual} to widen as the amount of available training data decreases. However, no such trend is observed. Across all dataset fractions, the two neural models exhibit nearly identical performance and even in the $\frac{1}{10000}$ dataset fraction where the standalone ADR model performs well, the baseline \textit{TSMixer} achieves lower error in most metrics.

These results do not support the common claim that incorporating physical knowledge improves sample efficiency, at least for the proposed architecture. Although the ADR model remains informative under severe data scarcity, \textit{TSMixer-ADRResidual} is unable to exploit this information to consistently improve its predictions. The limiting factor therefore appears to be not the quality of the underlying physical model itself, but rather the chosen mechanism for integrating its information into the neural network through additional physical features and physics-residual learning.

\section{Conclusion}

In this paper, we presented a physics-informed framework for PV power production forecasting that combines the ADR PV efficiency model with the TSMixer neural network architecture for time-series forecasting. We evaluated three commonly claimed advantages of PIML: faster convergence during training, improved generalization, and increased data efficiency. To this end, we designed three complementary experiments using a large-scale dataset comprising more than three million samples from over 3,700 PV installations distributed across Europe.

Across all experiments, our findings do not support these commonly reported benefits for the proposed forecasting framework. While the physics-informed models consistently achieved slightly lower validation loss during training, this advantage did not translate into improved performance on unseen test data. Similarly, neither the transferability nor the data-efficiency experiments demonstrated a systematic advantage of incorporating the ADR model into the neural network. Previous studies investigating PIML for PV forecasting have typically considered only one or a handful of installations and often relied on simpler ML models. In contrast, our evaluation combines a modern forecasting architecture with a highly heterogeneous continental-scale dataset. The results suggest that, under these conditions, the data-driven model is already capable of efficiently learning the relationships captured by the ADR model directly from the available data, leaving limited room for additional improvements through explicit physical modeling.

The data-efficiency experiment provides an additional important insight. Although the standalone ADR model remained relatively competitive when only very limited training data were available, the physics-informed \textit{TSMixer-ADRResidual} model was unable to consistently exploit this information to improve its predictions. Likewise, in the transferability experiments, the physics-informed model often generalized worse than the purely data-driven baseline when evaluated across different countries. Taken together, a plausible interpretation is that the additional physics-derived features only increase the model's ability to fit the training distribution without improving generalization. Since the ADR model introduces additional parameters fitted on historical data, the introduced physical representation may encode installation-specific biases absent in the data-driven baseline that do not transfer well to unseen conditions and actually cause the model to overfit the training data.

Our methodology also has several limitations that should be considered when interpreting the results. The ADR model is designed to estimate PV efficiency from accurate measurements of irradiance, module temperature, and installation geometry. In our dataset, however, none of these quantities is directly observed. Irradiance is obtained from day-ahead numerical weather prediction forecasts rather than measurements, module temperature is estimated using the Faiman thermal model, and installation metadata such as panel tilt and orientation are provided by system operators and may contain inaccuracies. Furthermore, the simplified physical model cannot account for installation-specific factors such as module aging, soiling, persistent shading, or snow cover. Consequently, inaccuracies in the physical model may partially offset the potential benefits of incorporating physical knowledge into the learning process. On the other hand, access to direct module temperature measurements or other detailed operational data is uncommon in practical forecasting applications, making our experimental setup representative of many real-world deployment scenarios.

Our results suggest that future work should focus less on incorporating physical models for their own sake and more on understanding what additional information they contribute beyond what modern neural forecasting models can already infer from the available data. Simply embedding a physical model into a neural architecture does not guarantee improvements in predictive accuracy, generalization, or data efficiency. More promising directions may include richer and more expressive physical models, alternative mechanisms for integrating physical knowledge into neural networks, or applications where the available data are insufficient for purely data-driven models to recover the underlying physical relationships.

% Finally, we believe that carefully designed negative results are an important part of scientific progress. Demonstrating where physics-informed machine learning does \emph{not} provide measurable benefits is just as valuable as reporting successful applications, helping to establish the conditions under which physical knowledge genuinely improves learning and where modern data-driven models are already sufficient.

\section*{Acknowledgements}

The work of Bernhard C. Geiger and Franz M. Rohrhofer was partially funded by the European Union’s HORIZON Research and Innovation Programme under grant agreement No 101120657, project ENFIELD (European Lighthouse to Manifest Trustworthy and Green AI).

This work of Peter Pavlík was partially funded by the European Union, under the project LorAI - Low Resource Artificial Intelligence, GA No. \href{https://doi.org/10.3030/101136646}{101136646}.

Research results were obtained using the computational resources procured in the national project National competence centre for high performance computing (project code: 311070AKF2) funded by European Regional Development Fund, EU Structural Funds Informatization of society, Operational Program Integrated Infrastructure.

Know Center is a COMET competence center that is financed by the Austrian Federal Ministry of Innovation, Mobility and Infrastructure (BMIMI), the Austrian Federal Ministry of Economy, Energy and Tourism (BMWET), the Province of Styria, the Steirische Wirtschaftsförderungsgesellschaft m.b.H. (SFG), the Vienna business agency and the Standortagentur Tirol. The COMET programme is managed by the Austrian Research Promotion Agency FFG.

\section*{Conflict of Interest}

The authors have no competing interests to declare that are relevant to the content of this article.

\section*{Computer Code Availability}
Training and evaluation source code is available at \href{https://github.com/kinit-sk/PV-PIML}{github.com/kinit-sk/PV-PIML}. The used dataset is available for non-commercial purposes upon request (requires license agreement with \href{https://pvoutput.org/}{PVOutput.org}).

\section*{Declaration of generative AI and AI-assisted technologies in the manuscript preparation process}

During the preparation of this work the authors used ChatGPT (GPT-5.5 Instant) to assist with language editing and manuscript refinement. After using this service, the authors reviewed and edited the content as needed and take full responsibility for the content of the published article.

% To print the credit authorship contribution details
\printcredits

%% Loading bibliography style file
%\bibliographystyle{model1-num-names}
\bibliographystyle{cas-model2-names}

% Loading bibliography database
\bibliography{cas-refs}

% Biography
%\bio{}
% Here goes the biography details.
%\endbio

%\bio{pic1}
% Here goes the biography details.
%\endbio

\end{document}